\relax
\documentclass[letterpaper]{article} 
\usepackage{aaai20}  
\usepackage{times}  
\usepackage{helvet}  
\usepackage{courier}  
\usepackage{url}  
\usepackage{graphicx}  
\usepackage{amsfonts}
\usepackage{courier}  
\usepackage{url}  
\usepackage{graphicx}  
\usepackage{amsfonts}
\usepackage{amssymb}
\usepackage{xspace}
\usepackage{booktabs}
\usepackage{amsmath,bm}
\usepackage{dsfont}
\usepackage{multirow}
\usepackage[autostyle, english = american]{csquotes}
\usepackage[ruled,linesnumbered]{algorithm2e}

\MakeOuterQuote{"}
\usepackage{ntheorem,lipsum} 
\theorembodyfont{\upshape} 
\newtheorem{definition}{Definition}

\usepackage[colorlinks=true, allcolors=blue]{hyperref}

\newcommand{\mname}{\text{Doctor2Vec}\xspace }
\begin{document}
%


\title{\mname: Dynamic Doctor Representation Learning for\\ Clinical Trial Recruitment}

\author{
\Large \textbf{Siddharth Biswal\textsuperscript{\rm 1,2}, Cao Xiao\textsuperscript{\rm 1}, Lucas M. Glass\textsuperscript{\rm 1}, Elizabeth Milkovits \textsuperscript{\rm 1}, and Jimeng Sun\textsuperscript{\rm 2}}\\
\textsuperscript{\rm 1} Analytic Center of Excellence, IQVIA, Cambridge, USA \\
\textsuperscript{\rm 2}  Computational Science and Engineering, Georgia Institute of Technology, Atlanta, USA 
}

\maketitle
\begin{abstract} 
Massive electronic health records (EHRs) enable the success of learning accurate patient representations to support various predictive health applications. In contrast, doctor representation was not well studied despite that doctors play pivotal roles in healthcare. How to construct the right doctor representations? How to use doctor representation to solve important health analytic problems?
In this work, we study the problem on {\it clinical trial recruitment}, which is about identifying the right doctors to help conduct the trials based on the trial description and patient EHR data of those doctors.
We propose \mname which simultaneously learns 1) doctor representations from EHR data and 2) trial representations from the description and categorical information about the trials. In particular, \mname utilizes a dynamic memory network where the doctor's experience with patients are stored in the memory bank and the network will dynamically assign weights based on the trial representation via an attention mechanism. Validated on large real-world trials and EHR data including 2,609 trials, 25K doctors and 430K patients, \mname demonstrated improved performance over the best baseline by up to $8.7\%$ in PR-AUC. 
We also demonstrated that the \mname embedding can be transferred to benefit data insufficiency settings including  trial recruitment in less populated/newly explored country with $13.7\%$ improvement  or for rare diseases with $8.1\%$ improvement in PR-AUC.




\end{abstract}

\section{Introduction}


The rapid growth of electronic health record (EHR) data and other health data enables the training of complex deep learning models to learn patient representations for disease diagnosis~\cite{choi2018mime,choi2016retain,doi:10.1093/jamia/ocy068}, risk prediction~\cite{10.1371/journal.pone.0195024}, patient subtyping~\cite{kdd2017subtyping,che2017rnn}, and medication recommendation~\cite{shang2018gamenet,Shang:2019:PGA:3367722.3367875}. However, almost all existing works focus on modeling patients. Deep neural networks for doctor representation learning are lacking.

Doctors play pivotal roles in connecting patients and treatments, including recruiting patients into clinical trials for drug development and treating and caring for their patients. Thus an effective doctor representation will better support a wider range of health analytic tasks. For example, identifying the right doctors to conduct the trials {\it site selection} so as to improve the chance of completion of the trials~\cite{hurtado2017improving}  and doctor recommendation for patients~\cite{8594889}. 

In this work, we focus on studying the  {\it clinical trial recruitment} problem using doctor representation learning.
Current standard practice calculates the median enrollment rate \footnote{Enrollment rate of a doctor is the number of patients enrolled by a doctor to the trial.} for a therapeutic area as the predicted enrollment success rate for all participating doctors, which is often inaccurate. In addition, some develop a multi-step manual matching process for site selection which is labor-intensive \cite{hurtado2017improving,potter2011site}. Recently, deep neural networks were applied on site selection tasks via static medical concept embedding using only frequent medical codes and simple term matching to trials~\cite{gligorijevic2019optimizing}. Despite the success, two challenges remain open.
\begin{enumerate}
    \item Existing works do not capture the time-evolving patterns of doctors experience and expertise encoded in EHR data of patients that the doctor have seen;
    \item Existing works learn a static doctor representation.  However, in practice given a trial for a particular disease, the doctor's experience of relevant diseases are more important. Hence the doctor representation should change based on the corresponding trial representation.
\end{enumerate}

To fill the gap, we propose \mname which simultaneously learns 1) doctor representations from longitudinal patient EHR data and 2) trial embedding from the multimodal trial description. In particular, \mname leverages a dynamic memory network where the representations of patients seen by the doctor are stored as memory while trial embedding serves as queries for retrieving from the memory.  \mname has the following contributions.
\begin{enumerate}
    \item \textbf{Patient embedding as a memory for dynamic doctor representation learning}. We represent doctors' evolving experience based on the representations from the doctors' patients.  The patient representations are stored as a memory for dynamic doctor representation extraction.
    \item \textbf{Trial embedding as a query for improved doctor selection}. We learn hierarchical clinical trial embedding where the unstructured trial descriptions were embedded using BERT \cite{devlin2018bert}. The trial embedding serves as queries of the memory network and will attend over patient representation and dynamically assign weights based on the relevance of doctor experience and trial representation to obtain the final context vector for an optimized doctor representation for a specific trial.
\end{enumerate}
We evaluated \mname using large scale real-world EHR and trial data for predicting trial enrollment rates of doctors. \mname demonstrated improved performance in site selection task over the best baselines by up to $8.7\%$ in PR-AUC. We also demonstrated that the \mname embedding can be transferred to benefit data insufficiency settings including trial recruitment in less populated/newly explored countries or for rare diseases. Experimental results show for the country transfer, \mname achieved  $13.7\%$ relative improvement in PR-AUC  over the best baseline. While for embedding transfer to rare disease trials, \mname achieved  $8.1\%$ relative improvement in PR-AUC over the best baseline.

\section{Related Works}

\subsubsection{Deep Patient Representation Learning} The collection of massive EHR data has motivated the use of deep learning for accurate patient representation learning and disease or risk prediction~\cite{doi:10.1093/jamia/ocy068,Fu:2019:PPL:3307339.3342159,kdd2017subtyping,choi2018mime}. In this work, we learn hierarchical patient representation in a similar way as ~\cite{choi2018mime}. But our focus is to construct doctor representation based on the embedding of their patients.

\subsubsection{Machine Learning Based Clinical Trial Recruitment}
Previously clinical trial enrollment either relies on simple statistics (e.g., medium enrollment) or manual matching~\cite{hurtado2017improving}. With the collection of clinical trial data, there has been some effort on developing machine learning-based models for trial site selection. For example,~\cite{van2017predicting} applied LR with  L1 regularization to determine a subset that is optimal for predicting site enrollment success. More recently, \cite{gligorijevic2019optimizing} learns static medical concept embedding and matches them to features derived from trial terms for site selection. However, no existing works learn trial embedding from multi-modal trial data and automatically match them to most relevant doctors.

\subsubsection{Memory Augmented Neural Networks} (MANN) have shown initial success in NLP research areas such as question answering ~\cite{DBLP:journals/corr/WestonCB14,sukhbaatar2015end,miller2016key,kumar2016ask}. Memory Networks ~\cite{DBLP:journals/corr/WestonCB14} and Differentiable Neural Computers (DNC)~\cite{graves2016hybrid} proposed to use external memory components to assist the deep neural networks in remembering and storing things. After that, various MANN based models have been proposed such as ~\cite{sukhbaatar2015end,kumar2016ask,miller2016key}. In healthcare, memory networks can be valuable due to their capacities in memorizing medical knowledge and patient history. DMNC~\cite{le2018dual} proposed a MANN model for medication combination recommendation task using EHR data alone. In ~\cite{shang2018gamenet}, the authors use a memory component to fuse multi-model graphs as a memory bank to facilitate medication recommendation.
\section{Method}
\begin{figure*}[h!]
\centering
    \includegraphics[width=0.8\textwidth]{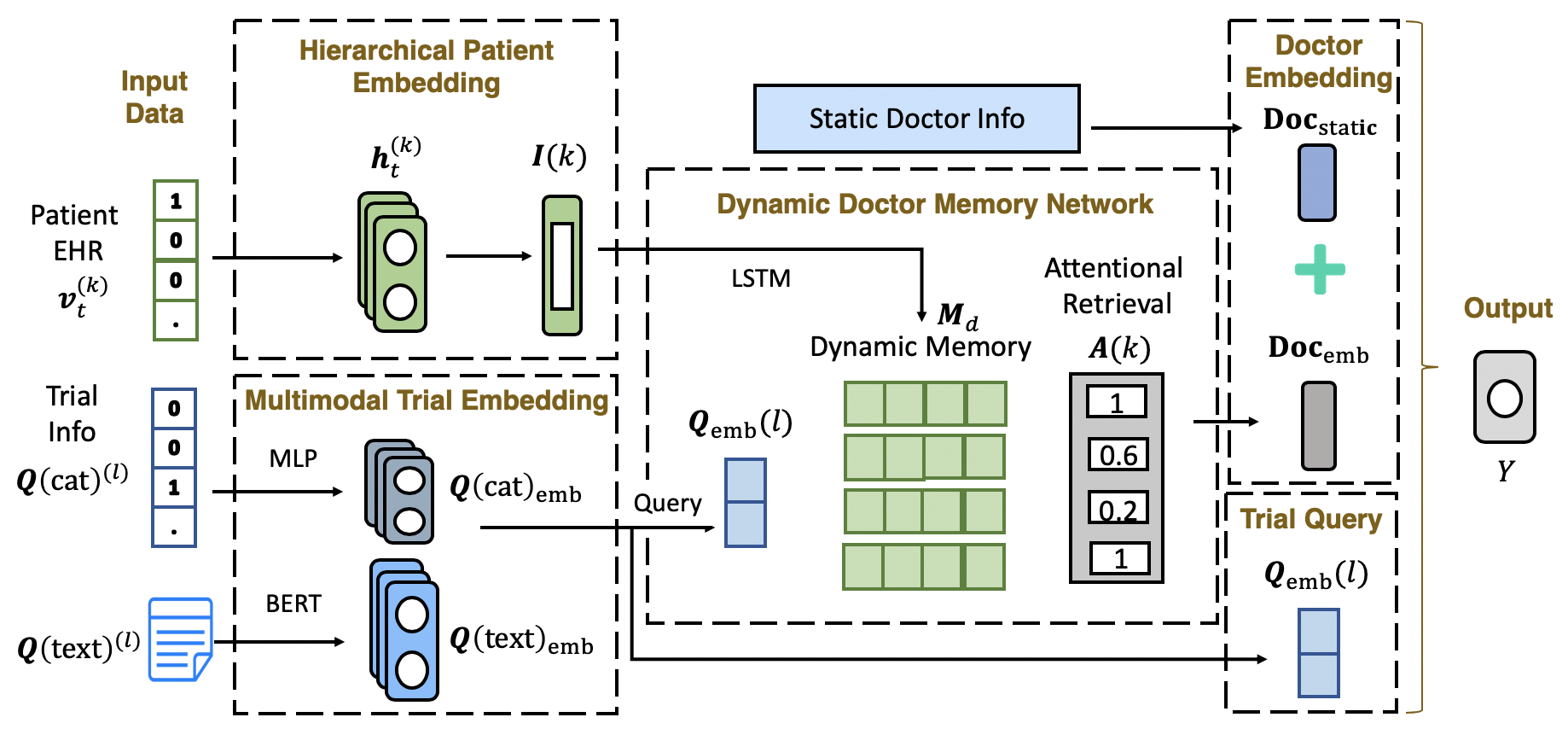} 
    \caption{The \mname Framework. (1) Hierarchical patient embedding: We obtain patient embeddings $\mathbf{h_t(k)}$ from patient visits using a Bi-LSTM with attention module. Unstructured text and categorical data are modeled using MLP and BERT respectively. (2) Multimodal clinical trial information embedding: The obtained clinical trial embeddings $\mathbf{Q_{emb}({l})}$ are fused together to form the query vector for the memory network.(3) Memory network module: This query vector to used to attend over the memory bank is obtained from combination of patient embeddings. The attentional vector is used to obtain the final doctor embedding . We combine the doctor embedding $\mathbf{Doc_{emb}}$ with clinical trial embedding $\mathbf{Q_{emb}(l)}$  and static information about the doctors $\mathbf{Doc_{static}}$  to predict the enrollment rate of the clinical trial (output).}
    \vskip -1em
    \label{fig:model_diagram}
\end{figure*}
\subsection{Problem Formulation}

\begin{definition}[Doctor Records] For each doctor, the clinical experience can be represented as a sequence of patients that the doctor has seen, denoted as $\mathbf{C(m)} = \{\mathbf{P_{1}^{(m)}},\mathbf{P_{2}^{(m)}},\cdot, \mathbf{P_{k}^{(m)}}\}$ where $m$ indicates the $m$-th doctor. 
Here each patient can also be represented as a sequence of multivariate observations $\mathbf{P(k)} = \{\mathbf{v_{1}^{(k)}},\mathbf{v_{2}^{(k)}},\cdot, \mathbf{v_{T}^{(k)}}\}$ where $k$ indicates the $k$-th patient and subscript $1,2,\ldots, T$ indicates different visits for the $k$-th patient. Each visit $\mathbf{v}_t^{(k)}$ is the combination of diagnosis codes $\mathbf{c_d}$, medication codes $\mathbf{c_m}$, and procedure codes $\mathbf{c_p}$. Medical codes $\mathbf{c_d}, \mathbf{c_p}, \mathbf{c_m} \in {0,1}^{|C|}$ are represented as multi-hot vectors and ${|C|}$ represent the total size of the code sets. We also use the demographic information available about doctors and patients available. These static information are denoted as $\mathbf{Doc_{static}}$ which are used with other features.  
\end{definition}

\begin{definition}[Clinical Trial Data] Clinical trial data comprises of two data modalities: the trial descriptions in unstructured text and the categorical features such as trial phase,  primary indication, primary outcome, secondary outcome, and study type.
We denote each clinical trial as a combination of text data and categorical data. 
\begin{align*}
    & \mathbf{Q(cat)(l)}=\{\mathbf{f_{1}^{(l)}},\mathbf{f_{2}^{(l)}},\cdot, \mathbf{f_{v}^{(l)}} \}\\
    & \mathbf{Q(l)} = [\mathbf{Q(cat)(l)}; \mathbf{Q(text)(l)}] 
\end{align*}
where $\mathbf{l} \in \{1,2,\cdot,\cdot, L\}$ is the index of clinical trials, $\mathbf{{f}_i}$ is the representation for the categorical trial features, and $\mathbf{v}$ is the number of categorical features in a trial.
\end{definition}

\begin{table}[!hb]
\centering
  \caption{Notations used in \mname. }
  \label{basic_symbols}
  \begin{tabular}{cl}
    \toprule
    Symbol & Definition and description \\
    \midrule
    $\mathcal{D}$ & Set of unique diagnosis codes\\
    $\mathcal{P}$ & Set of unique procedure codes\\
    $\mathcal{M}$ & Set of unique medication codes\\
    $\mathbf{C(m)}$ & Notation for a doctor \\
    $\mathbf{P_{k}^{(m)}}$ & Notation for a patient for m-th doctor \\
    $\mathbf{v_{t}^{(k)}}$ & the $t$-th visit of Patient $k$\\
    $ \mathbf{Q(l)}$ & The $\mathbf{l}$-th clinical trial \\
    $\mathbf{Q(cat)^{(l)}}$ & Categorical features of the $\mathbf{l}$-th clinical trial \\
    $\mathbf{Q(text)^{(l)}}$ & Text features of the l-th clinical trial \\
    $\mathbf{I_{f}(k)}$ &  Final patient representation  \\ 
    $\mathbf{Doc_{emb}}$ &  Vector representation of a Doctor\\
    $\mathbf{Doc_{static}}$ & Static features of a Doctor\\
  \bottomrule
\end{tabular}
\end{table}

\subsection{ \mname Framework}
As illustrated in Fig. 1, \mname includes the following components: \textit{hierarchical patient embedding},  \textit{multimodal clinical trial information embedding}, a \textit{memory network module} for dynamical doctor information retrieval, and a memory output module for generating the current time doctor representation. This doctor representation is used with clinical trial representation to predict the \textit{enrollment rate} of the clinical trial. Next, we will first introduce these modules and then provide details of training and inference.

\subsubsection{Hierarchical Patient Embedding} As discussed before, a doctor has seen a set of patients during his/her medical practice. We model that the doctor's experience and expertise by the function over the embeddings of their patients' EHR data. If a patient has seen two doctors, the corresponding portion of the EHR data will be modeled as two separate patients for both doctors, respectively. 

To learn patient representation, motivated by ~\cite{choi2018mime}, we leverage the inherent multilevel structure of EHR data to learn patient embedding hierarchically. The hierarchical structure includes the patients on the top, followed by visits the patient experiences over time, then at the leaf level the set of diagnosis, procedure, and medication codes recorded for each visit. Here given clinical visits of a patient, denoted as $\mathbf{v_t^{(k)}}$, we firstly pass these visits through a multi-layer perception (MLP) to get visit embedding $\mathbf{h_t^{(k)}}$ as follows.
\begin{equation}
        \mathbf{h_t^{(k)}} = \mathbf{W_{emb}}\mathbf{v_t^{(k)}}    
\label{eqn:MLP}
\end{equation}
Without ambiguity, we will ignore the patient index $(k)$ for brevity.
Next to learn a patient embedding based on a sequence of visit embeddings, we pass visit embedding $\mathbf{h_k}$ to a bi-directional long short-term memory (bi-LSTM) networks and then add an attention layer on top of the bi-LSTM to attend on important visits. 

\begin{align}
            &\mathbf{g_1,g_2,... g_t} = \text{bi-LSTM}(\mathbf{h_1, h_2,..., h_t})\\
            &{e_t} = \mathbf{w_\alpha^T g_t}+{b_{\alpha}}\\
            &{\alpha_1,\alpha_2, ..\alpha_t} = \mathrm{\mathrm{softmax}({e_1,e_2, ..e_t})}
\end{align}
Then we obtain final patient context vector $\mathbf{I(k)}$ by summing over attended visit representations as given by Eq.~\ref{eqn:p_emb}.
\begin{equation}
\mathbf{I(k)}= \sum \mathbf{\alpha_t} \cdot \mathbf{h_t}
\label{eqn:p_emb}
\end{equation}
Here $\mathbf{I(1)} \ldots \mathbf{I(k)}$ are patient representations that will be fed into dynamic doctor memory bank as  memory vectors.

\subsubsection{Multimodal Trial Information Embedding}

Clinical Trials are conducted to evaluate a specific drug or procedure. We use public and private information about clinical trials. Here we obtained clinical trial descriptions from publicly available clinical trial database clinicaltrials.gov. The collected trial description comprises multiple data modalities, including unstructured text and categorical features. In this module, we employ a multi-modal method to embed them in a shared space.
\begin{enumerate}
    \item \textbf{Unstructured Text}.
Each trial has a text description for inclusion and exclusion criteria which describe the requirements and restrictions for  recruiting to the trial.  For unstructured text, we applied the BERT~\cite{devlin2018bert} model. BERT builds on the Transformer~\cite{vaswani2017attention} architecture and improves the pre-training using a masked language model for bidirectional representation. Essentially BERT model is a stack of Transformer blocks. Each transformer block is a combination of self-attention block and feedforward layer. \textbf{Pretraining}: We use the same pre-training techniques as in \cite{devlin2018bert}  (1) Masked Language Model: This modeling task consists of masking $15\%$ of the input tokens and using the model to predict the masked tokens. This is trained with cross entropy loss. (2) Next sentence prediction task: In this task two sentences are fed to BERT. The model outputs a binary prediction of whether these two sentences are in consecutive order. The final pre-training objective function based on the two tasks is the sum of the log-likelihood of the masked tokens and the log-likelihood of the binary variable indicating whether two sentences are consecutive. We pretrained a  BERT model using MIMIC text data~\cite{johnson2017mimic} to extract embeddings for each word and combine the word embeddings to obtain the final text embeddings. In order to pretrain BERT on text corpus, we first obtain preprocessed data as required by BERT which includes tokenization, indexing and masking. The procedure is formulated as below. Denote unstructured text associated with each trial as  $\mathbf{Q(text)}^{(l)}$, the embedding $\mathbf{Q(text)_{emb}}$ is given by Eq.~\ref{ref:trial_embed_text}.  We average over embeddings obtained from each word to compute the final embedding for the entire document.


\begin{align}
 \mathbf{Q(text)_{emb}} = \mathrm{BERT}(\mathbf{Q(text)}^{(l)})
\label{ref:trial_embed_text}
\end{align}


\item \textbf{Categorical Features}.
The categorical features of each clinical trial include the geographic location of the trial, hospital system, primary and secondary therapeutic area, pharmaceutical company information, phase of the trial, condition or disease, objectives of the trial, intervention model, etc. More details of the categorical features in the Appendix. The dimensions of these categorical variables ranged from $18$ to $1456$. For these features, we first encode them using one-hot vectors and then pass the one-hot vectors through multi-layer perception (MLP) layer. This can be expressed as below. Denote the categorical features as $\mathbf{Q(cat)}^{(l)}$, the categorical feature embedding is obtained as in Eq.~\ref{ref:trial_embed_cat}.
\begin{align}
    \mathbf{Q(cat)_{emb}}= \mathbf{W_c} \mathbf{Q(cat)}^{(l)}+ \mathbf{b_c}
\label{ref:trial_embed_cat}
\end{align}

\end{enumerate}

After obtaining embeddings from both types of clinical trial data, we fuse the embeddings from categorical data $\mathbf{Q(cat)_{emb}}$ and text data $\mathbf{Q(text)_{emb}}$ to obtain the final embedding $\mathbf{Q_{emb}}(l)$. We fuse these two embeddings weighted multiplicative fashion as in Eq.~\ref{ref:trial_embed}.

\begin{align}
 \mathbf{Q_{emb}}(l)= \mathbf{(\mathbf{W_{ci}}\mathbf{Q(cat)_{emb}}+ \mathbf{b_{ci}})} \odot \nonumber \\ (\mathbf{W_{ti}}\mathbf{Q(text)_{emb}}+\mathbf{b_{ti}})
\label{ref:trial_embed}
\end{align}
where $\odot$ is element-wise multiplication.
The clinical trial embeddings $\mathbf{Q_{emb}}(l)$ will be fed into dynamic doctor memory network as the query to extract related patient representation memory.  

\subsubsection{Dynamic Doctor Memory Networks}
Since each doctor sees a diverse set of patients, doctor representation should be dynamically constructed for a given trial as opposed to a static embedding vector staying the same for all trials. The way we achieved that is by a dynamic memory network where patients are stored as memory vectors of the doctor. Then using a trial embedding as a query, we fetch the relevant patient vectors from the memory bank and dynamically assemble a doctor representation for this trial. 

Inspired by ~\cite{DBLP:journals/corr/WestonCB14}, four memory components \textbf{I}, \textbf{G}, \textbf{O}, \textbf{R} are proposed which mimics the architecture of modern computer in storing and processing information.
\begin{enumerate}
    \item \textbf{Input Memory Representation}.
    This layer converts the patient representations to the input representation. We pass all the patient representations through a dense layer to obtain the input representations.
        \begin{equation}
            \mathbf{I_f(k)}=\mathbf{W_i}\mathbf{I(k)}+ \mathbf{b_i}
        \end{equation}
    \item \textbf{Generalization}.
    Typically generalization can be referred to as the process of updating memory representation for the memory bank. In our case, we use the patient representations to initialize the memory representation $\mathbf{M_{d}}$ which is the combination of all the patient representations. We then apply an LSTM layer to update the memory via multiple iterations. 
    \begin{equation}
        \mathbf{M_{d}}= \mathrm{LSTM}(\mathbf{I_f(1)},\cdots,\mathbf{I_f(k)})
    \end{equation}

\item \textbf{Output}.
  In this step, the final output memory representation is generated. We calculate the relevance between trial embedding $\mathbf{Q_{emb}(l)}$ and doctor embedding  $\mathbf{M_{d}}$ to obtain  ${{A(k)}}$ as the  attention vector over patient representations.
    \begin{equation}
        {{A(k)}} = \mathrm{softmax}[\mathbf{Q_{emb}(l)^T} \mathbf{M_{d}}]
    \end{equation}
\item \textbf{Response}. In this step, we obtain the final $\mathbf{Doc_{emb}}$ using the patient embeddings and attention weights over the patients.
\begin{equation}
    \mathbf{Doc_{emb}}= \sum {A(k)}\mathbf{I_f(k)}
\label{eqn:context_rec}
\end{equation}

\end{enumerate}

We use the doctor representation which is composed of patients and the clinical trial representation to obtain a final context vector.
Besides dynamic doctor embedding $\mathbf{Doc_{emb}}$, we also include static information about doctors in the final embedding such as their educational history, length of practice, length of practice into the feature vector. The resulting final embedding vectors are then fed into a fully connected layer and passed through a softmax to obtain class labels. 
\begin{equation}
    \mathbf{Y}= \mathrm{Softmax}([\mathbf{Doc_{emb}};\mathbf{Q_{emb}(l)}; \mathbf{Doc_{static}}])
\label{eqn:final_pred}
\end{equation}
where the input to Softmax are concatenation of dynamic doctor embedding $\mathbf{Doc_{emb}}$, trial query embedding  $\mathbf{Q_{emb}(l)}$
and static doctor embedding $\mathbf{Doc_{static}}$.

The enrollment rate category is obtained by binning the continuous enrollment rate. We divide the continuous enrollment scores into five discrete classes ranging at $0\sim0.2$, $0.2\sim0.4$, $0.4\sim0.6$. $0.6\sim0.8$, $0.8\sim1.0$. The 5 enrollment categories are used labels for classification.

\subsection{Training and Inference}

During training,  We train our models by minimizing the cross entropy loss to optimize $\mathbf{W_{emb}}$, $\mathbf{W_c}$, $\mathbf{W_i}$, weight matrices of  Bi-LSTM. We denote the network parameters by $\theta_c$ which is updated by optimizing for the loss function.

\begin{equation}
     \mathcal{L} = -\frac{1}{N}\sum(\mathbf{y_i} log(\hat{\mathbf{y_i}}) +(1-\mathbf{y_i})^T log(1-\hat{\mathbf{y_i}}))
\label{eqn:loss}
\end{equation}

In the inference phase, we use calibrated threshold value for obtaining predicted labels from the predicted probability values where we obtain the probability values from the final layer of the network as mentioned in Eq. \ref{eqn:final_pred}. Our \mname model is summarized in Algorithm~\ref{alg:model_training}.

\begin{algorithm}[!htb]
\caption{Model Training for \mname }\label{alg:model_training}
\KwIn{Training dataset, input $(\mathbf{C},\mathbf{Q})$ and target $\mathbf{Y}$; epochs $N_{epoch}$} 
\KwOut{Trained model for enrollment rate prediction with parameter $\mathbf{\theta_c}$}
Initialization\; 
\For{$i=1,\dots,N_{epoch}$} 
{
    \ForEach{mini-batch in the training set}
    {
        Obtain $\mathbf{h_k(m)}$ using MLP in Eq. \ref{eqn:MLP} \;
        Compute $\mathbf{I(k)}$ using BiLSTM and Attn. Eq. \ref{eqn:p_emb}\;
        Compute $\mathbf{Q_{emb}}$ using combination of MLP and BERT by trial embedding; \\
        $\mathbf{I_f(k)}$ is obtained from $\mathbf{I(k)}$ ;\\
        $\mathbf{A(k)}$ is generated from inner product of $\mathbf{Q_{emb}}$ and $\mathbf{I_f(k)}$  \\
        Compute Doctor representation $\mathbf{Doc_{emb}}$ in Eq. \ref{eqn:context_rec};\\
        Combine $\mathbf{Doc_{emb}},\mathbf{Q_{emb}(l)} \mathbf{Doc_{static}}$  for the final prediction Eq. \ref{eqn:final_pred};\\
        Calculate prediction loss $\mathcal{L}$ using Eq. \ref{eqn:loss};;\\
        Update parameters according to the gradient of     $\mathcal{L}$;\\
    
    }
}
\end{algorithm}

\section{Experiment}

We designed experiments to answer the following questions.

\noindent \textbf{Q1}: Does \mname have better performance in predicting clinical trial enrollment to support site selection?


\noindent \textbf{Q2}: Can \mname embedding perform in transfer learning
setting for trials across countries or across diseases?

\subsubsection{Implementation} We implemented \mname~\footnote{Code: https://github.com/sidsearch/Doctor2vec} with PyTorch 1.0~\cite{pytorch}. For training the model, we used Adam~\cite{adamKingmaB14} with the mini-batch of 128 samples. The training was performed on a machine equipped with an Ubuntu 16.04 with 128GB memory and Nvidia Tesla P100 GPU.

\subsubsection{Data Source} 
We obtained patient and trial information from the following three data sources. 
\begin{enumerate}
    \item We extracted trial data from IQVIA's real-world patient and clinical trial database, which can be accessed by request~\footnote{https://www.iqvia.com/insights/the-iqvia-institute}. It contains 2609 clinical trials formed during 2014 and 2019. This dataset includes 25894 doctors across 28 countries. It includes both unstructured eligibility criteria and categorical features including the geographic location of the trial, hospital system, primary, secondary therapeutic areas, drug names, etc. The data also includes outcome measures such as the trial enrollment rate. In ground truth, the distribution of the enrollment categories are 12\%, 33\%, 37\%, 12\%, 6\% respectively for $0\sim0.2$, $0.2\sim0.4$, $0.4\sim0.6$. $0.6\sim0.8$, $0.8\sim1.0$ bins of enrollment score.  
    \item We also obtained real world patient claims dataset from Database 1. This dataset contains a longitudinal treatment history from 430,239 patients over 7 years. In addition to medical codes about diagnosis, procedure, medication, it also includes information about doctors such as specialty, education, hospital location, geographical location.
    \item We also extract clinical trial descriptions from publicly available clinical trial database clinicaltrials.gov. We match the trial information with our Database 1 on NCT ID which is a universal clinical trial ID.
\end{enumerate}

\subsubsection{Enrollment Rate} Enrollment rate for each investigator is defined as 
\begin{align*}
        \text{Enrollment Rate} = \frac{\text{\# sub. randomized}-\text{\# sub. discontinued}}{\text{enrollment window}}
\end{align*}
    
After obtaining the enrollment numbers, we perform a min-max normalization step to obtain normalized enrollment rate which is between 0 and 1. In this normalization step, we only consider the investigators associated with each clinical trial.

\begin{table}[h]
\centering
\caption{Data Statistics}
\label{tab:stats}
\begin{tabular}{lccc}
\toprule
\# of clinical trials   & 2,609     \\ 
\# of doctors         & 25,894    \\ 
\# of doctor-trial pair(samples)         & 102,487    \\
\# of patients & 430,239  \\ 
\hline
Avg \# of Dx codes per visit & 4.23  \\
Max \# of Dx codes per visit & 56 \\
Avg \# of Procedure codes per visit & 1.23 \\
Max \# of Procedure codes per visit & 18 \\
Avg \# of Med codes per visit & 9.36 \\
\hline
\end{tabular}
\end{table}

\subsubsection{Baselines}
We consider the following baselines.
\begin{enumerate}
    \item Median Enrollment (Median). Current industry standard that considers the median enrollment rate for each  therapeutic area as estimated rate for all trials in that area.
    \item Logistic Regression (LR).
    We combine the medication,diagnosis and procedure codes along with the clinical trial information to create feature vectors, and then apply LR to predict the enrollment rate category. 
    \item Random Forest (RF)~\cite{Breiman:2001:RF:570181.570182}.
    We combine the medication, diagnosis and procedure codes along with the clinical trial information to create feature vectors and then pass it to RF to predict the enrollment rate category.
    \item AdaBoost~\cite{Schapire:1999:BIB:1624312.1624417}. 
    We combine the medication,diagnosis and procedure codes along with the clinical trial information to create feature vectors and then apply AdaBoost classifier to predict the enrollment rate categories.
    \item Multi-layer Perceptron (MLP).
    We use MLP to process doctor features. In this case, we obtain the doctor features by converting all the visit vectors associated with a doctor to a count vector of different diagnosis, medication, procedure codes. We convert categorical information of clinical trials to multi-hot vectors and obtain TF-IDF features from text information of clinical trials.
    \item Long Short-Term Memory Networks (LSTM)~\cite{hochreiter1997long}.
    We process all the temporally ordered visit vectors associated with a doctor using an LSTM. The embedding obtained from LSTM is concatenated with embedding obtained from categorical and text information of clinical trials to predict enrollment rate.
    \item DeepMatch~\cite{gligorijevic2019optimizing}
    In this model, the features for the doctors are obtained by collecting the top 50 most frequent medical codes and passed through an MLP layer to obtain an embedding vector. This embedding is concatenated with embedding obtained from categorical and text information of clinical trials via MLP and TF-IDF to finally predict enrollment rate. 
\end{enumerate}

\subsubsection{Evaluation Metrics}
To evaluate the performance of enrollment prediction , We used   PR-AUC as the metric for the classification task, and the coefficient of determination ($R^2$) score for the regression task.  Details of the metrics are provided in appendix.

\subsubsection{Experiment Setup and Evaluation Strategies}
We split our data into train, test, validation split with 70:20:10 ratio. We also ensured that the clinical trails are unique and no overlap in train, test, validation split. We used Adam~\cite{kingma2014adam} optimizer at learning rate 0.001 with learning rate decay. We fix the best model on evaluation set within 200 epochs and report the performance in test set. Details about reproducibility including hyperparameters are provided in Appendix.

\subsection{Q1: \mname achieved the best predictive performance in clinical trial enrollment prediction}
We conducted experiments for both classification (e.g., predict enrollment rate category) and regression (e.g., predict actual rate) tasks.   Results are provided in Table~\ref{tab:regression_results}. From the results, we  observe that  \mname achieved the best performance in both settings.


\begin{table}[h!]
\centering
\caption{\mname achieves the best performance on both metrics  in predicting actual enrollment rate (regression task)  and rate categories (classification task) compared to state-of-the-art baselines. Results of ten independent runs.}
\label{tab:regression_results}
\begin{tabular}{lcccc}
\toprule
                     & PR-AUC &  $R^2$ Score  \\ \hline
Median              & $0.571\pm0.014$   &  $0.54\pm0.072$    \\ 
LR                & $0.672\pm0.041$  &  $0.314\pm0.082$   \\ 
RF                  & $0.731\pm0.034$  &  $0.618\pm0.034$    \\
AdaBoost             & $0.747\pm0.002$ &  $0.684\pm0.146$ \\
MLP                  & $0.761\pm0.019$ &  $0.762\pm0.049$  \\
LSTM                  &$0.792\pm0.034$ & $0.780\pm0.621$  \\
DeepMatch            & $0.735\pm0.068$ & $0.821\pm0.073$ \\
\textbf{\mname}         & $\boldsymbol{0.861\pm0.021}$ &  $\boldsymbol {0.841\pm0.072}$ \\ \bottomrule
\end{tabular}
\end{table}

For category classification, \mname has $8.7\%$ relative improvement in PR-AUC  over the best baseline LSTM. Among the baselines, the Median method performs the worst, indicating the population level information is not accurate enough for each individual trial. Tree based models such as RF and Adaboost performs better than Median enrollment and LR, which can be attributed to the large number of features they leverage as well as their ability of distilling complex features. MLP performs better than tree based models due to having adequate number of layers for better capturing information and ability control over-fitting. The LSTM network further improves over MLP since it is able to extract  the temporal information present from the visits of patients. Compared with these approaches, the DeepMatch models achieved much lower PR-AUC since the model leverages the 50 most frequent codes for medical concept embedding, thus missing many important information of the doctors.

In actual rate prediction task, \mname gains  $2.4\%$ relative improvement in $R^2$  over best baseline DeepMatch. As for the baselines, the LR model performs the worst, indicating linear models cannot capture the complex and temporal information in the data. Median enrollment is worse than most baselines but better than LR since for some more common diseases median enrollment can be a good predictor. Again, MLP and LSTM work better than tree-based models due to they can better capture complex features.
 DeepMatch in the regression settings tends to perform better than MLP and LSTM, which can be attributed to the majority of actual scores being in the range of [0.4-0.65] leading to improved performance.

\subsection{Q2: \mname can perform well in trial recruitment prediction even across countries and across diseases}

One major challenge for clinical trial recruitment is when conducting trials in a less populated country or a country that is newly explored, or building a trial for a rare disease, the recruitment rate is often hard to estimate since there is not enough historical data to refer to. In this section we design two experiments to explore whether the embedding learned by \mname will be useful in order to benefit the aforementioned data insufficiency settings.
\begin{enumerate}
    \item Trained on United States data and transfer to a less populated/newly explored country;
    \item Trained on common diseases and transfer to rare/low prevalence diseases.
\end{enumerate}

For the first experiment, we trained \mname on $1443$ clinical trials in the United states during the time 2014-2019  and test on 47 clinical trials in South Africa during the time 2014-2019. We perform the same model transfer for all baselines. Results are provided in Table~\ref{tab:knowledge_transfer}. 

For the second experiment, we test the model on $38$ clinical trials for drugs about idiopathic pulmonary fibrosis (IPF, a rare lung disease ) and inflammatory bowel disease(IBD, a low prevalence chronic inflammatory bowel disease). The model was trained on $2569$ clinical trials from the rest of the available diseases.  We perform the same model transfer for all baselines. Results are provided in Table ~\ref{tab:knowledge_transfer_rare_disease}.

\begin{table}[h!]
\centering
\caption{\mname achieves the best performance when we transfer the model trained on  US data to predict trial enrollment in South Africa.}
\label{tab:knowledge_transfer}
\begin{tabular}{lcccc}
\toprule
                         & PR-AUC   & $R^2$ Score  \\ \hline
Median                   &   $0.524\pm0.032$        &    $0.420\pm0.039$        \\
LR 	                    & $0.601\pm0.023$        &    $0.279\pm0.014$             \\
RF   	                & $0.661\pm0.038$         &  $0.552\pm0.048$   	\\
AdaBoost                & $0.672\pm0.01$         &   $0.581\pm0.039$     	\\
LSTM                     &  $0.758\pm0.013$       &  $0.721\pm0.025$  \\ 
DeepMatch              &  $0.703\pm0.087$      &    $0.756\pm0.031$  \\ 
\textbf{\mname }       &  $ \boldsymbol{0.862\pm0.003}$       &    $\boldsymbol{0.817\pm0.025}$   \\
\hline
\end{tabular}
\end{table}

\begin{table}[h!]
\centering
\caption{\mname achieves the best performance when we transfer the model trained on common disease trials to rare and low prevalence disease trials.}
\label{tab:knowledge_transfer_rare_disease}
\begin{tabular}{lcccc}
\toprule
                                                         & PR-AUC   & $R^2$ Score  \\ \hline
Median                          &   $0.413\pm0.013$        &   $0.387\pm0.001$         \\
LR 	                         &  $0.521\pm0.021$        &   $0.225\pm0.028$            \\
RF   	                         &  $0.610\pm0.019$         &   $0.517\pm0.032$         	\\
AdaBoost                      &  $0.623\pm0.002$         &   $0.548\pm0.046$          	\\
LSTM                            &  $0.725\pm0.002$       &     $0.623\pm0.038$     \\ 
DeepMatch                   &  $0.638\pm0.021$      &      $0.678\pm0.049$   \\ 
\textbf{\mname }            &  $\boldsymbol{0.784\pm0.032}$       &   $\boldsymbol{0.716\pm0.014}$   \\
\hline
\end{tabular}
\end{table}

For both settings,  \mname  performs much better than state-of-the-art baselines.  For the country transfer, \mname achieved  $13.7\%$ relative improvement in PR-AUC over best baseline LSTM and $8.1\%$ relative improvement in $R^2$ over best baseline DeepMatch.
While for embedding transfer to rare disease trials, \mname achieved  $8.1\%$ relative improvement in PR-AUC  over the best baseline LSTM and $5.6\%$ relative improvement in $R^2$ over best baseline DeepMatch.


For country transfer, we also examine the $R^2$ scores. Based on the $R^2$ values, the DeepMatch model and LSTM model accounts for $69.2\%$ and $67.3\%$ of the variance in the data, respectively. While our \mname  accounts for $83.6\%$ of the variance. This shows our model and the prediction fit more to the real observation.

\subsection{Case Study}
We present case studies to demonstrate the effectiveness of the proposed \mname model. \\

\subsubsection{Phase I  trial for Gemcitabine plus Cisplatin}  This phase I trial is a combination  cancer therapy.  A doctor in the US who has worked in internal medicine during the past 3 years has run the trial. The actual enrollment rate is $0.72$. The rate estimation provided by the best baseline LSTM is $0.57$. While the estimated rate from \mname is $0.69$, which is much closer to the ground truth. The reason for \mname to perform more accurately is the internal medicine doctor has a broader coverage of diseases. Baseline models consider all these diseases that the doctor treated when measuring the match between the doctor and the trial.
While \mname was able to focus more on the patients who had cancer diagnosis instead of all patients which leads to improved prediction.


\subsubsection{Phase II trial for Alzheimer's Disase}
This phase II trial is about an amyloid drug for treating Alzheimer's patients. A doctor in the US who has treated cancer during the past 4 years runs the trial and has a trial enrollment rate at $0.62$. The estimated rate from the best baseline LSTM model is $0.45$. \mname predicts the enrollment rate will be $0.58$, which is much closer to the ground truth. 
For this case, \mname is more accurate because \mname is able to learn better doctor representations for this doctor by focusing on the neurological disease patients compared to other disease type patients seen by the doctor.

\section{Conclusion}
In this work, we proposed \mname, a doctor representation learning based on both patient representations from longitudinal patient EHR data and trial embedding from the multimodal trial description.  \mname leverages a dynamic memory network where the representations of patients seen by the doctor are stored as memory while trial embedding serves as queries for retrieving the memory. Evaluated on real world patient and trial data, we demonstrated via
trial enrollment prediction tasks that \mname can learn accurate doctor embeddings and greatly outperform state-of-the-art baselines. We also show by additional experiments that the \mname embedding can also be transferred to benefit the data insufficient setting (e.g., model transfer to less populated/newly explored country or from common disease to rare disease) that is highly valuable yet extremely challenging for clinical trials.

\section*{Acknowledgement}
This work was in part supported by the National Science Foundation award IIS-1418511, CCF-1533768 and IIS-1838042, the National Institute of Health award NIH R01 1R01NS107291-01 and R56HL138415.


\bibliographystyle{aaai}
\bibliography{sample}

\clearpage

\section{Appendix}

\subsection{Data Preprocessing}

\begin{itemize}
    \item \textbf{Investigator Data}:
     We group different treatments done by investigators over time based on time. This creates a temporal view of the treatments performed by the investigator over time. The temporal view of the investigator consists of all the diagnoses, procedures, medication codes prescribed to patients.
    We combine the diagnosis, procedure, medication codes into a single treatment. These codes are further converted to multi-hot representations which are used for the input for the model. Other information such as physician's location, specialty area, professional certifications, etc is also used for the input for the model. We provide the details of these features in the supplementary section.
    
    \item\textbf{Clinical Trials Data}: 
    Clinical trials data were preprocessed to have combine information about the trial from both private and public sources. Our private dataset contains information about the success of the clinical trials. It also contains information about the pairing of investigators with the trials information. Similar to the investigator data, there are lot of variables which are categorical in nature and some variables are textual in nature. We convert the categorical variables into multi-hot representation after removing null values. We combine the public information about the trials with the private information using the a global ID. This helps us get more context about the clinical trials. The textual information is converted continuous value embeddings using BERT(Bidirectional Encoder Representations from Transformers). We combine all these information to provide as the input for the clinical trial embedding module. 
    We will provide the details about the different variables in the supplementary section of this manuscript.

\end{itemize}

\subsection{Notations}
\begin{itemize}
    \item Definition 1(Patients):
    We denote the patients as 
    \begin{equation}
         P(n) = {x_{1}^{(n)},x_{2}^{(n)},... x_{T}^{(n)}}
    \end{equation}
   where n $\in$ {1,2,....N} patients in the cohort and T is the number of visits by individual patients.
   
   Each visit $x_{1}^{N}$ is the combination of diagnosis, medication, procedure codes ${c_d, c_p, c_m}$.

    \item Definition 2(Providers):

    \item Definition 3(Clinical Trials):

\end{itemize}

\subsection{Enrollment Rate Definition}

Enrollment rate is defined as 
\begin{align}
    \text{Enrollment Rate} = \frac{\text{subjects randomized}-\text{subjects discontinued}}{\text{enrollment window}}
\end{align}

\subsection{Baselines}

\subsection{Evaluation Metrics}

\begin{enumerate}
\item Precision: Here precision measures the fraction of relevant doctors among the recommended doctors.
\begin{align*}
    \mathrm{Precision} = \frac{\text{TP}}{\text{TP}+\text{FP}}
\end{align*}
\item Recall: Here recall measures the fraction of relevant doctors that have been recommended over the total amount of qualified doctors.
\begin{align*}
  \mathrm{Recall} = \frac{\text{TP}}{\text{TP}+\text{FN}}
\end{align*}
\item PR-AUC: Area Under the Precision-Recall Curve.
    \begin{equation*}
        \text{PR-AUC} = \sum_{k = 1}^{n} \mathrm{Prec}(k) \Delta \mathrm{Rec}(k), 
    \end{equation*}
    where $k$ is the $k$-th precision and recall operating point ($\mathrm{Prec}(k), \mathrm{Rec}(k)$).
\end{enumerate}

\subsection{\mname Model Hyper parameter details}
In this section, we provide details about model hyperparameters:

\textbf{Bi-LSTM}:
\begin{itemize}
    \item Activation function:tanh
    \item bias usage: True
    \item Number of hidden units: 124
    \item kernel Regularizer: L2 regularizer
\end{itemize}

\textbf{MLP for categorical variables of clinical data}:
\begin{itemize}
    \item Number of layers: 4
    \item Dimension of layers: [128,256,128,64]
\end{itemize}

\textbf{MLP in memory network}:
\begin{itemize}
    \item Number of layers: 3
    \item Dimension of layers: [128,128,64]
\end{itemize}

\subsection{BERT background and formulation}

Here we provide more details about the BERT formulation used in our method. We also provide details about pretraining performed. BERT model is based on a multi-layer Transformer encoder \cite{vaswani2017attention} and it is usually BERT is pre-trained using two unsupervised tasks (1) Masked Language Model and (2) Next Sentence Prediction.

\textbf{Input formatting}: Since BERT is trained on specific type of symbols, we need to format our input data in that format to be used with the pretrained model. We use special tokens to mark the beginning ([CLS]) and separation/end of sentences ([SEP]). This also requires token IDs from BERT’s tokenizer and mask IDs to indicate which elements in the sequence are tokens and which are padding elements. BERT uses WordPiece tokenization method which splits words into multiple splits.

\textbf{Transformer as BERT building block}:
The transformer model was proposed in Attention is all you need paper \cite{vaswani2017attention}. 

\textbf{BERT model definition}:
BERT is basically a trained Transformer Encoder stack. BERT model has large number of encoder layers or Transformer blocks.  Each transformer block contains a self attention block with feed forward blocks. Each encoder block has a residual connection around it. It is also followed by a layer-normalization step.

\textbf{Pretraining for BERT}:
Pretraining of BERT is performed by two tasks (1) Masked LM (2) Next sentence prediction 

\textbf{Word embeddings from BERT}:
Pre-trained BERT can be used to create contextualized word embeddings. It has been shown that BERT based word embeddings are highly successful in downstream tasks. We extract the vectors from the last four layers of the BERT model. These vectors are of $768$ dimension. Instead of concatenating these vectors, we averaged these vectors to obtain the final word embeddings which are of $768$ dimension. 

\subsection{Baseline Hyperparameter and Implementation Details}
The parameters are initialized as per the original paper. The model dimensionality is set to 768. We use the Adam optimizer  with a learning rate of 0.0002 . The maximum sequence length supported by the model is set to 512, and the model is first trained using shorter sequences. The model is trained using a maximum sequence length of 128 for 75,000 iterations on the masked language modeling and next sentence prediction tasks, with a batch size 32.The model is trained on longer sequences of maximum length 512 for an additional 75,000 steps with a batch size of 4.

\begin{itemize}

    \item \textbf{Logistic Regression Hyperparameters}
    \begin{itemize}
        \item Penalty function: L2
        \item tolerance(stopping criteria): 1e-5
        \item Inverse regularization strength: 1.2
        \item Bias(fit intercept): True
        \item class weight: Balanced
        \item Solver: liblinear
        \item maximum iteration: 3000
    \end{itemize}
    
    \item \textbf{Random Forest Hyperparameters}
    \begin{itemize}
        \item number of estimators: 12
        \item criterion: Gini impurity
        \item Maximum Depth: 16
        \item Minimum samples split: 5
        \item Minimum samples leaf: 3
        \item Minimum weight fraction leaf: 0
        \item Maximum Features: automatic selection
        \item Maximum leaf nodes: 4
    \end{itemize}
    
    \item \textbf{Adaboost Hyperparamters}
    \begin{itemize}
        \item Booster: gbtree
        \item learning rate: 0.2
        \item Max Depth: 8
        \item Minimum Child Weight: 1
        \item Maximum Delta Step:0.4
        \item lambda(L2 regularization): 0.3
    \end{itemize}
    
    \item \textbf{MLP Hyperparameters}
    \begin{itemize}
        \item Activation Function: ReLU
        \item Number of Layers: 6
        \item Size of Layers: [512, 512, 512, 256, 128, 64]
        \item Optimizer: Adam
        \item Learning Rate: 0.001
    \end{itemize}

    \item \textbf{RNN Hyperparameters}
       \begin{itemize}
        \item Activation Function: tanh
        \item Number of layers: 2
        \item Size of Layers: [128, 128]
        \item Optimizer: Adam
        \item Learning Rate: 0.001
    \end{itemize}
    \item \textbf{DeepMatch Hyperparameters}
       \begin{itemize}
        \item Medical concept embedding dense layer dimension: 200
        \item Medical concept embedding activation: ReLU
        \item Clinical trial embedding layer dimension: 300
        \item Optimizer: Adam
        \item Learning Rate: 0.001
    \end{itemize}

\end{itemize}

\subsection{BERT improvement compared to word2vec}

\begin{table}[!htb]
\centering
\caption{Text embedding method comparison}
\label{tab:word_emb_comp}
\begin{tabular}{lcccc}
\toprule
                  & PR-AUC  & F1-score  \\ \hline
Word2vec    &  $0.801\pm0.018$ & $0.812\pm0.024$  \\ 
char-CNN    & $0.819\pm0.024$  & $0.823\pm0.038$  \\ 
BERT        &  $0.861\pm0.021$ & $0.841\pm0.032$  \\
\hline
\end{tabular}
\end{table} 

\subsection{MSE results for overall enrollment prediction}
We have reported the mean squared error rates for the regression task here. Due to space constraint, this results is not presented in table 3.

\begin{table}[h!]
\centering
\caption{\mname achieves the best performance in terms of MSE metrics for the regression task}
\label{tab:regression_results}
\begin{tabular}{lcccc}
\toprule
                     & MSE &   \\ \hline
Median              & $1.220\pm0.041$       \\ 
LR                  & $28.210\pm1.341$    \\ 
RF                  & $0.981\pm0.032$      \\
AdaBoost             & $0.736\pm0.015$  \\
MLP                  & $0.663\pm0.148$   \\
LSTM                  &$0.462\pm0.047$  \\
DeepMatch            & $0.381\pm0.023$\\
\textbf{\mname}     & $\boldsymbol{0.221\pm0.038}$  \\ \bottomrule
\end{tabular}
\end{table}

\end{document}